\documentclass[10pt,twocolumn,letterpaper]{article}

\usepackage{cvpr}              % To produce the CAMERA-READY version
\usepackage{graphicx}
\usepackage{amsmath}
\usepackage{amssymb}
\usepackage{booktabs}
\usepackage{indentfirst}

\usepackage[pagebackref,breaklinks,colorlinks]{hyperref}

\usepackage[capitalize]{cleveref}
\crefname{section}{Sec.}{Secs.}
\Crefname{section}{Section}{Sections}
\Crefname{table}{Table}{Tables}
\crefname{table}{Tab.}{Tabs.}

\def\cvprPaperID{*****} % *** Enter the CVPR Paper ID here
\def\confName{CVPR}
\def\confYear{2022}

\begin{document}

\title{Billet Number Recognition Based on Test-Time Adaptation }

\author{
  Yuan Wei and Xiuzhuang Zhou$^*$\\
}

\maketitle

\renewcommand{\thefootnote}{}
\footnotetext{$^*$Corresponding author. The authors are with the School of Intelligent Engineering and Automation, Beijing University of Posts and Telecommunications.}

%%%%%%%%% ABSTRACT
\begin{abstract}
   
During the steel billet production process, it is essential to recognize machine-printed or manually written billet numbers on moving billets in real-time. To address the issue of low recognition accuracy for existing scene text recognition methods, caused by factors such as image distortions and distribution differences between training and test data, we propose a billet number recognition method that integrates test-time adaptation with prior knowledge. First, we introduce a test-time adaptation method into a model that uses the DB network for text detection and the SVTR network for text recognition. By minimizing the model’s entropy during the testing phase, the model can adapt to the distribution of test data without the need for supervised fine-tuning. Second, we leverage the billet number encoding rules as prior knowledge to assess the validity of each recognition result. Invalid results, 
which do not comply with the encoding rules, are replaced. Finally, we introduce a validation mechanism into the CTC algorithm using prior knowledge to address its limitations in recognizing damaged characters. Experimental results on real datasets, including both machine-printed billet numbers and handwritten billet numbers, show significant improvements in evaluation metrics, validating the effectiveness of the proposed method.
\end{abstract}

%%%%%%%%% BODY TEXT
\section{Introduction}
\label{sec:intro}

Steel billets are widely used in industries such as construction, military, and automotive manufacturing. During production, essential information—such as company ID, furnace number, sequence number, production date, and steel grade—is encoded into billet numbers according to predefined rules. These identifiers are applied to the billet surface either via machine printing or manual handwriting. By providing a unique identifier for each billet, factories can efficiently cross-check production plans, enabling precise tracking and streamlined subsequent processing. The current practice of recording billet numbers relies primarily on manual transcription, which requires significant manpower to sustain shift work across production lines. The unpredictable timing of billet arrivals forces workers to remain on-site for extended periods, resulting in high workloads and low efficiency. Prolonged shifts often lead to fatigue-induced transcription errors, undermining operational accuracy. Furthermore, the harsh steel production environment, characterized by excessive noise and high temperatures, exacerbates worker fatigue and poses serious health risks.

In recent years, advancements in computer vision have enabled the efficient resolution of many low-skill, repetitive tasks through algorithmic approaches. Optical Character Recognition (OCR)\cite{1}, one of the most mature fields in computer vision, can be applied to the recognition of billet numbers. There are already several well-established and efficient pre-trained scene text recognition models available\cite{21}\cite{22}\cite{23}\cite{24}. However, in real-world production settings, the factory environment introduces significant distribution differences between pre-trained models' training data and actual billet number data. For deep learning models, this often results in substantial performance degradation. Additionally, obtaining a large volume of annotated billet data for model fine-tuning or training is challenging. Furthermore, due to factors like high temperatures and oxidation, billet numbers often exhibit distortions, peeling, or damage. Most mainstream scene text recognition models employ the Connectionist Temporal Classification (CTC) \cite{2}algorithm for post-processing. However, we observed that CTC has limitations when dealing with damaged characters, as it tends to produce omissions during recognition. These issues make it difficult to deploy existing pre-trained scene text recognition models in production scenarios, as their performance in such environments is often unsatisfactory.

To address these challenges and improve the accuracy of billet number recognition, this paper proposes the following solutions:
(1) Incorporating test-time adaptation (TTA) into pre-trained scene text recognition models: TTA refines model parameters in an unsupervised manner during testing, enabling pre-trained models to adapt to actual test data. This approach improves the accuracy of scene text recognition models for billet number recognition tasks without requiring original training data or modifications to the original training process.
(2) Integrating encoding rules as prior knowledge: Billet numbers adhere to certain encoding rules. We organized these rules and designed a reasoning method based on this prior knowledge to automatically correct recognition results that do not conform to the encoding rules.
(3) Enhancing the handling of CTC's ``Blank" tokens: By leveraging prior knowledge, we classified the ``blank" tokens introduced by the CTC algorithm and restored a portion of these tokens to their actual recognition results. This adjustment addresses CTC's limitations in handling damaged characters and improves recognition accuracy.

%------------------------------------------------------------------------
\section{Related Works}
\label{sec:formatting}
%-------------------------------------------------------------------------
\subsection{Billet Number Recognition }

Current research on billet number recognition can be broadly divided into two categories: traditional image processing methods and deep learning-based approaches.

In traditional image processing methods, Zhang et al. \cite{3} designed a billet number recognition system by investigating the image characteristics of billet numbers. They proposed a binarization algorithm that combined the Otsu method and Niblack thresholding to mitigate the effects of lighting on billet number recognition. The connected domain method was employed for image segmentation, followed by neural network-based character recognition. While this binarization method effectively reduces the impact of uneven lighting, it fails to completely separate the background from the foreground and struggles to handle connected characters. Zhou et al.\cite{4} proposed an improved template matching algorithm for billet character recognition. Although template matching—a traditional character recognition method—performs well when characters are intact and exhibit no significant deformation, its accuracy declines significantly in the presence of such disturbances. To address issues like connected characters, Dong et al.\cite{5} introduced a multi-level projection segmentation algorithm and employed a recursive segmentation approach to locate billet number regions. However, this method demands precise imaging performance, requiring the target to remain in a specific position within the image. Wu et al.\cite{6}  developed an automatic billet number recognition algorithm based on support vector machines to meet the automation requirements of steel mills. While this method achieves high classification accuracy, its performance heavily relies on the selection of character features. 

With the advancements in deep learning, OCR technology has matured and become an active research topic in areas such as license plate recognition, image retrieval \cite{7}, and industrial automation. Koo et al.\cite{8} trained convolutional neural network (CNN) models to directly detect or recognize steel plate identification numbers. Zhao et al.\cite{9} developed a recognition classification model based on character encoding rules and recognition confidence, categorizing the results into correct, suspicious, and erroneous classes. Suspicious and erroneous results were manually verified. Although this method improved accuracy, it did not fully automate the process. Sun et al.\cite{10} built a handwritten billet number recognition network using YOLO.\  Building on this, Wang et al.\cite{11}proposed a method combining YOLOv5 with ESRGAN, reconstructing low-resolution input images into high-resolution output images for subsequent recognition. Ge et al.\cite{12} used a line-scan camera to continuously capture steel plate images and proposed a linear dot-matrix character recognition method for plate surfaces, achieving recognition of high-frequency dot-matrix characters on steel plates. Xu et al.\cite{13} introduced a recognition network based on flow alignment and attention mechanisms, incorporating a hybrid dilation convolution block to expand the receptive field. This enhanced the network's ability to effectively capture scale features. Additionally, they constructed flow features and distortion maps to enable upsampling, addressing information loss during feature fusion.

Although these methods achieved promising results, they primarily focused on improving the accuracy of character image recognition, without addressing performance degradation caused by factors such as the complex environments in steel mills, online recognition of moving billets, and varying billet image distributions.

%-------------------------------------------------------------------------
\subsection{Test-Time Adaptation}

Our work involves test-time adaptation (TTA) or test-time training (TTT) to enhance the generalizability of models.

Test-time training (TTT) adapts models during the testing phase by introducing a self-supervised auxiliary task. For example, Wang et al.\cite{14} proposed self-consistency, while Sun et al.\cite{15} introduced rotation prediction as auxiliary tasks. The choice of auxiliary tasks significantly impacts the model's adaptability, and they must be trained in the same way during the model’s training phase.

On the other hand, test-time adaptation (TTA) aims to adapt models during the testing phase without requiring access to training data or modifications to the training process. For instance, DUA\cite{16}, utilizes a small portion of test data and its augmentations to adapt BatchNorm statistics. TENT\cite{17}employs entropy minimization to fine-tune the BatchNorm layers during testing. Niu et al. proposed SAR\cite{18}, which excludes high-gradient samples to promote weight flattening. Lee et al. developed PL\cite{19}, which uses reliable pseudo-labels to fine-tune parameters. SHOT \cite{20} combines entropy minimization with pseudo-labeling techniques.

%------------------------------------------------------------------------
\section{Method}

The overall framework of the algorithm is shown in Figure 1. The DB network is employed as the text detection network, while the SVTR network serves as the text recognition network to obtain the initial recognition results. During this process, the test-time adaptation module is utilized to update the model parameters dynamically. Subsequently, the recognition results are corrected using prior knowledge through the CTC algorithm and further refined during the post-processing stage. 

\begin{figure}[h]
  \centering
   \includegraphics[width=0.8\linewidth]{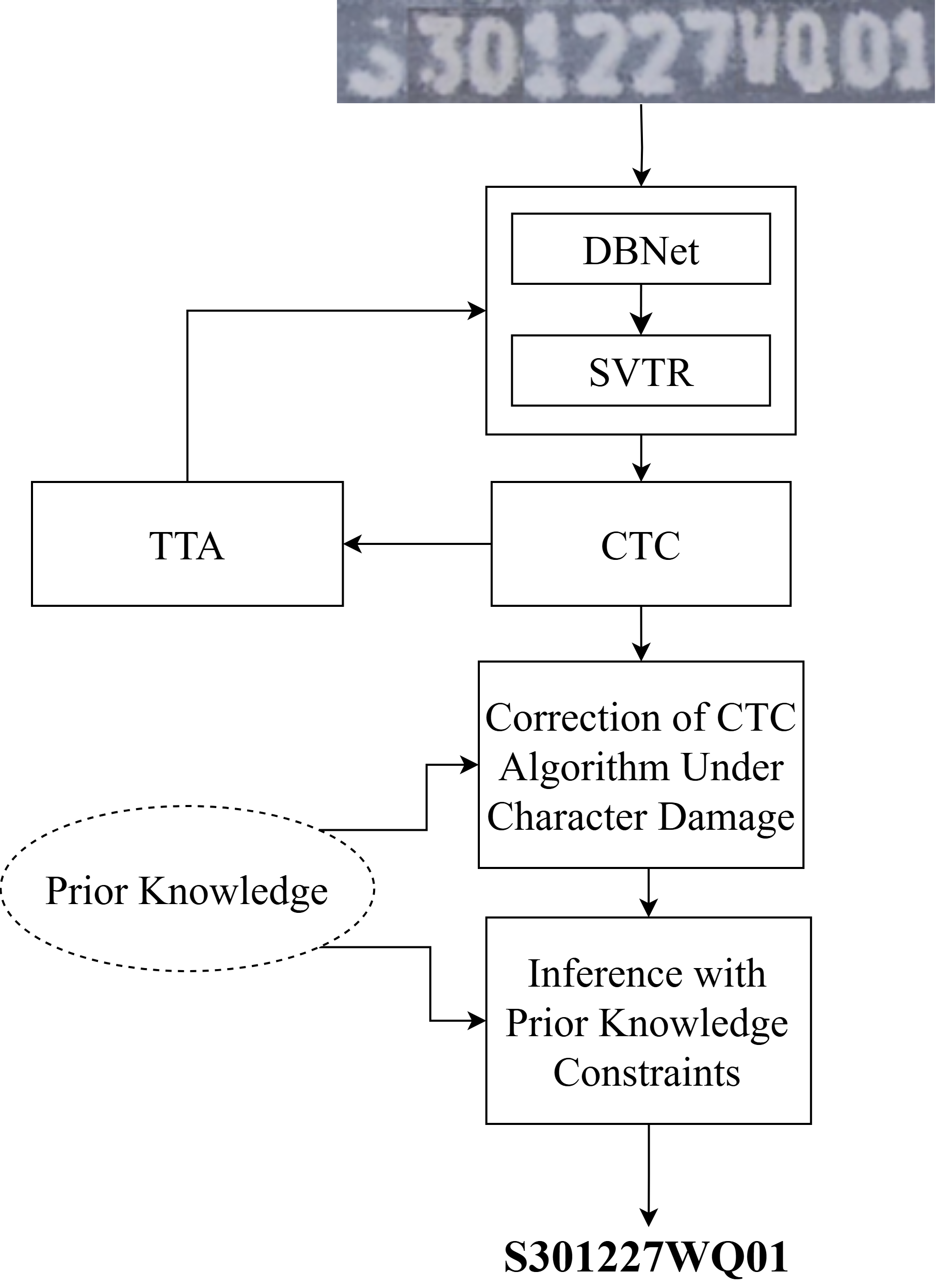}

   \caption{Overview of our method.}
   \label{fig:onecol}
\end{figure}

\subsection{Text Detection and Recognition Networks}

The steel billet number recognition task can be regarded as a specific case of scene text recognition, which is typically divided into two stages: text detection and text recognition. In the text detection phase, we utilize the DB (Differentiable Binarization) network\cite{21}, a segmentation-based text detection algorithm. The core of this network lies in its differentiable thresholding mechanism, which binarizes the image dynamically to distinguish text regions from the background. The DB network achieves high-precision text detection through the following steps: First, a residual convolutional neural network (ResNet) is used to extract features from the input image, generating multi-scale feature maps. Next, a Feature Pyramid Network (FPN) fuses feature maps of different scales, enabling the network to capture text information of various sizes. Then, the fused feature maps are passed into a DB module, which produces two outputs: a probability map and a threshold map. In the probability map, each pixel value indicates the likelihood of belonging to a text region. The cross-entropy loss function is used to supervise the training process, assigning a label of 1 to pixels in text regions and 0 to those outside. In the threshold map, each pixel value represents the binarization threshold for that pixel. An L1 loss is employed to minimize the difference between the network-predicted threshold and the ground truth, which is computed based on the statistical characteristics of text information in the neighboring pixel. The differentiable thresholding function, as shown in Eq.(1), is applied to the probability map $P$ and threshold map $T$ pixel-wise to generate the final binarized image. $k$ denotes a gain factor. The binarized image is supervised using a cross-entropy loss with ground-truth labels. 

\begin{equation}
  \hat{B}=\frac{1}{1+e^{-k\left(P_{i,j}-T_{i,j}\right)}}
  \label{eq:important}
\end{equation}

In the text recognition stage, the SVTR\cite{22} network is utilized, with its structure illustrated in Figure 2. First, the input image is processed through a patch embedding module, consisting of two cascaded convolutional blocks, to split it into character components. Each component corresponds to a small section of text characters in the image. Next, the features pass through three stages, which progressively downsample in the height dimension. Every stage comprises two operations, including mixing, merging, or combining, to extract features at different scales. The extracted features are then passed through a fully connected layer for linear prediction, producing the character sequence. SVTR captures multi-granular character features, enabling it to detect local details within individual characters while simultaneously modeling long-range global dependencies between characters.

\begin{figure}[h]
  \centering
   \includegraphics[width=0.9\linewidth]{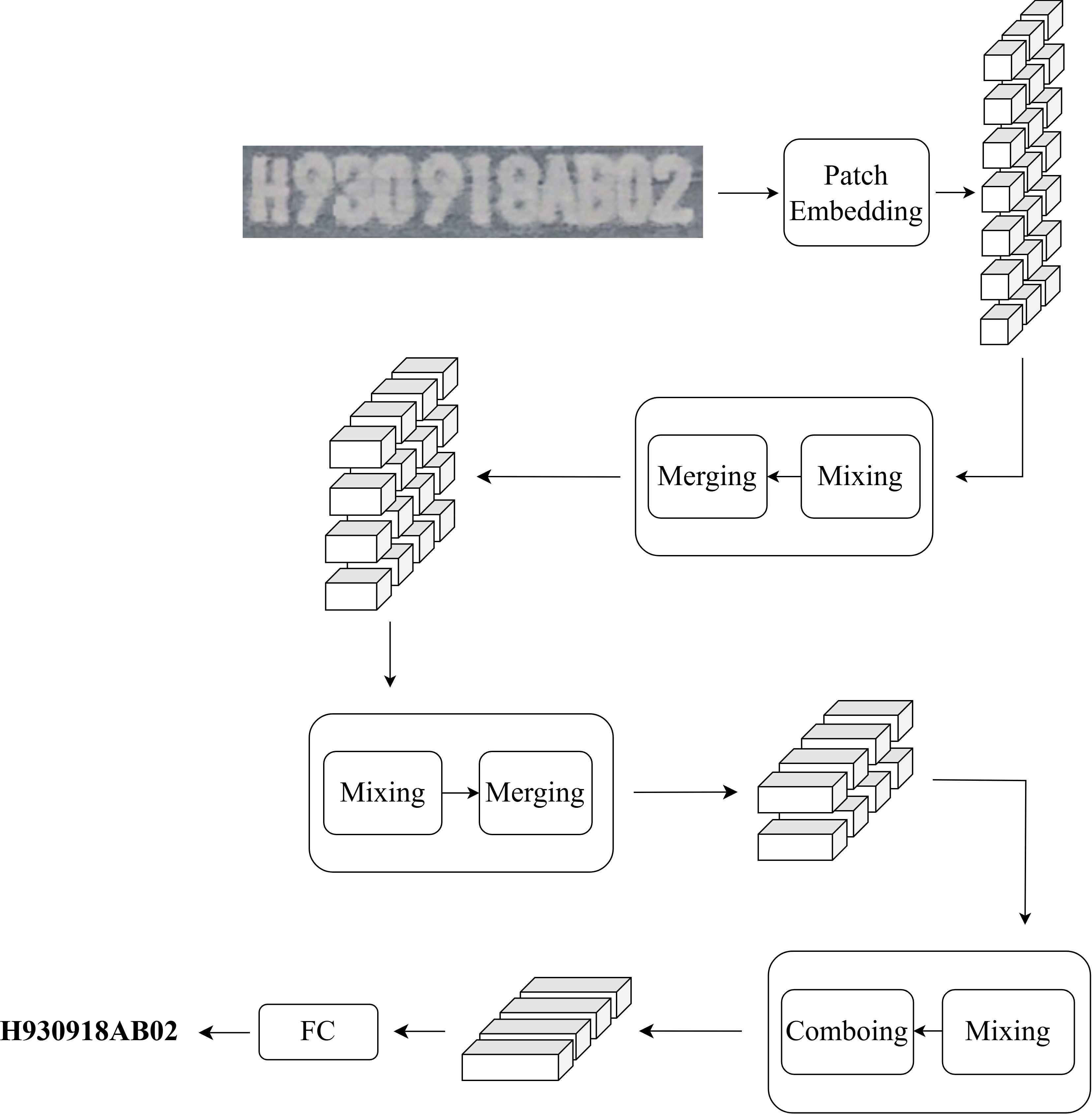}
   \caption{SVTR Network Architecture.}
   \label{fig:onecol}
\end{figure}

\subsection{Test-time Adaptation Method}

Traditional machine learning techniques typically learn from a large amount of source domain data $ x^s $ and corresponding labels $ y^s $, resulting in a model $ f_\theta $ with parameters $ \ \theta $. Afterward, the model is fixed for inference. This paradigm often performs exceptionally well when the test and training data share the same distribution. However, in practical applications, the distribution of target domain data $ x^t $ frequently deviates from the distribution of source domain training data $ x^s $, a phenomenon known as domain shift, which degrades the model's inference performance.

Unlike the above methods, test-time adaptation relies solely on the test data $x^t$. It uses self-supervised or unsupervised approaches to fine-tune the model during testing and then employs the updated model to make final predictions. This process depends only on the model's predictions and does not require labels for the test data, nor does it need source domain data or modifications to the training process. Thus, it is a highly convenient method for addressing domain shifts.

For model $\hat{y}=f_\theta\left(x^t\right)$, the model parameters $\theta$ are obtained by training on the data from the source domain $\ x^s$ and labels $\ y^s$. It has been observed that for the test samples in the target domain, the higher the image entropy, the higher the prediction error rate of the model, as shown in Figure 3. Therefore, entropy can serve as a measure of the model’s performance on the current dataset. Thus, the chosen optimization objective during the test phase is to minimize the model's entropy: $H\left(\hat{y}\right)=-\sum_{c}{p({\hat{y}}_c)logp({\hat{y}}_c)}$
where $p({\hat{y}}_c)$ represents the probability that the model predicts class $c$. This is an unsupervised metric that relies solely on the model’s predictions. Since the model adapts to the data in an unsupervised manner during testing, it is required that the model has been trained on source domain data in a supervised manner and that its output provides the probability values for each class. This requirement aligns perfectly with the basic configuration of the SVTR network.

\begin{figure}[h]
  \centering
   \includegraphics[width=1\linewidth]{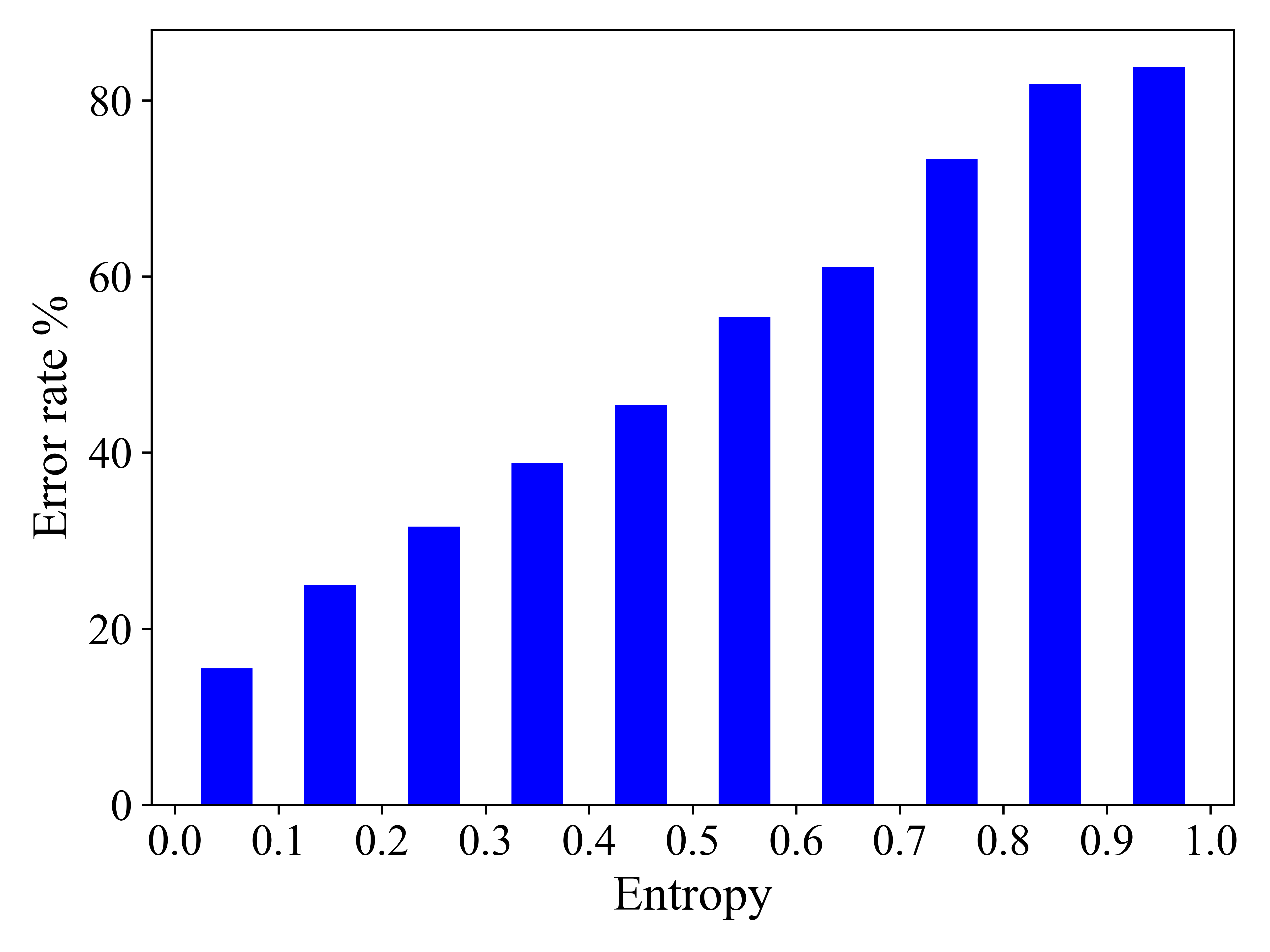}
   \caption{The relationship between entropy and error rate.}
   \label{fig:onecol}
\end{figure}

During the testing phase, we need to modify the model parameters $\ \theta$. However, the number of model parameters is typically enormous. Making extensive modifications may result in the model losing its original recognition capability and consume significant time, which fails to meet industrial real-time requirements. Considering the above, we adopt the approach proposed by Tent et al.\cite{17}, which updates only certain parameters in the batch normalization (BN) layers during the testing phase. The batch normalization layer consists of four components: mean $\ \mu$, standard deviation $\sigma$, scaling parameter $\gamma$, and shift parameter $\ \beta$. Among these, $\ \mu$ and $\sigma$ are obtained based on statistical data: \( \mu \leftarrow E[x^t], \sigma^2 \leftarrow E[(\mu - x^t)^2] \). We update the remaining affine transformation parameters $\gamma$, $\beta$ by minimizing the entropy: \( \gamma \leftarrow \gamma + \frac{\partial H}{\partial \gamma}, \beta \leftarrow \beta + \frac{\partial H}{\partial \beta} \).

During the testing phase, we fix all model parameters except $\gamma$ and $\ \beta\ $. During forward propagation, the current batch's $\ \mu$ and $\ \sigma$ are computed, and during backward propagation, $\gamma$ and $\ \beta\ $ are updated by minimizing $H\left(\hat{y}\right)$. In practical industrial scenarios, as the test data arrives continuously, the model can update itself incrementally, becoming increasingly adapted to the distribution of the current test data.

\begin{table*}[h] % 使用 table* 环境
  \centering
  \caption{The data and the structure of the loss function used by different adaptation method.}
  \label{tab:twocolumn}
  \begin{tabular}{@{}lcccc@{}}
    \toprule
    Method & Source domain data (\%) & Target domain data (s) & Train loss & Test loss\\
    \midrule
    Fine-tuning  & -- & $x^t,\ y^t$ & $L(x^t,\ y^t)$ & --\\
    Domain adaptation & $x^s,\ y^s$ & $x^t$ & $L\left(x^s,\ y^s\right)+L(x^t,\ y^t)$ & --\\
    Test-time training& $x^s,\ y^s$ & $x^t$ & $L\left(x^s,\ y^s\right)+L(x^s)$ & $L(x^t)$\\
    Test-time adaptation& -- & $x^t$ & -- & $L(x^t)$\\
    \bottomrule
  \end{tabular}
\end{table*}

\subsection{Inference with Prior Knowledge Constraints}
\subsubsection{Automatic Correction of Recognition Results Based on Encoding Rules}

\indent  The steel billet number adheres to specific encoding rules. A complete steel billet number consists of several parts, including the company name, production date, furnace number, serial number, connector, and sequence number. Each part must comprise exclusively letters or digits. These encoding rules can be utilized as prior knowledge to guide the post-processing of recognition results. Specifically, during the recognition process, the model stores the probabilities of all potential recognition results for each character position in a dictionary and sorts them. If the result with the highest probability conforms to the encoding rules, it is selected as the recognition result for that position. If it does not conform, the second most probable result is chosen, and the process continues until a valid result is identified. For example, the first character, representing the company name, must be a letter. If the result with the highest probability is a digit, it clearly violates the encoding rules. The algorithm will then proceed to evaluate the second most probable result and so on until a valid result is found.

\subsubsection{Correction of CTC Algorithm Under Character Damage}

After obtaining the initial recognition results from the SVTR network, the Connectionist Temporal Classification (CTC) algorithm is applied for post-processing. CTC is a commonly used post-processing algorithm in text recognition tasks. Since the exact text length is unknown beforehand and there is often spacing between characters, the number of recognition boxes typically exceeds the actual number of characters in the text. Some recognition boxes cannot align with the characters to be recognized, resulting in unavoidable redundant information in the recognition output. The core idea of the CTC algorithm is to address the matching problem between input and output sequences. Specifically, CTC introduces a ``blank" label (denoted as ``\_" in the following) into the original dictionary. If the algorithm detects that the character in a recognition box is incomplete or unaligned, it assigns the label ``\_" to represent redundant information. After recognizing all the detection boxes, the algorithm performs two operations on the recognition results: 1. Remove consecutive duplicate characters. 2. Remove the ``\_" labels. For example, if the original recognition result is B\_63\_6\_0\_21\_B\_B\_06\_, the CTC-processed result becomes B636021BB06, as shown in Figure 4.

\begin{figure}[h]
  \centering
   \includegraphics[width=1\linewidth]{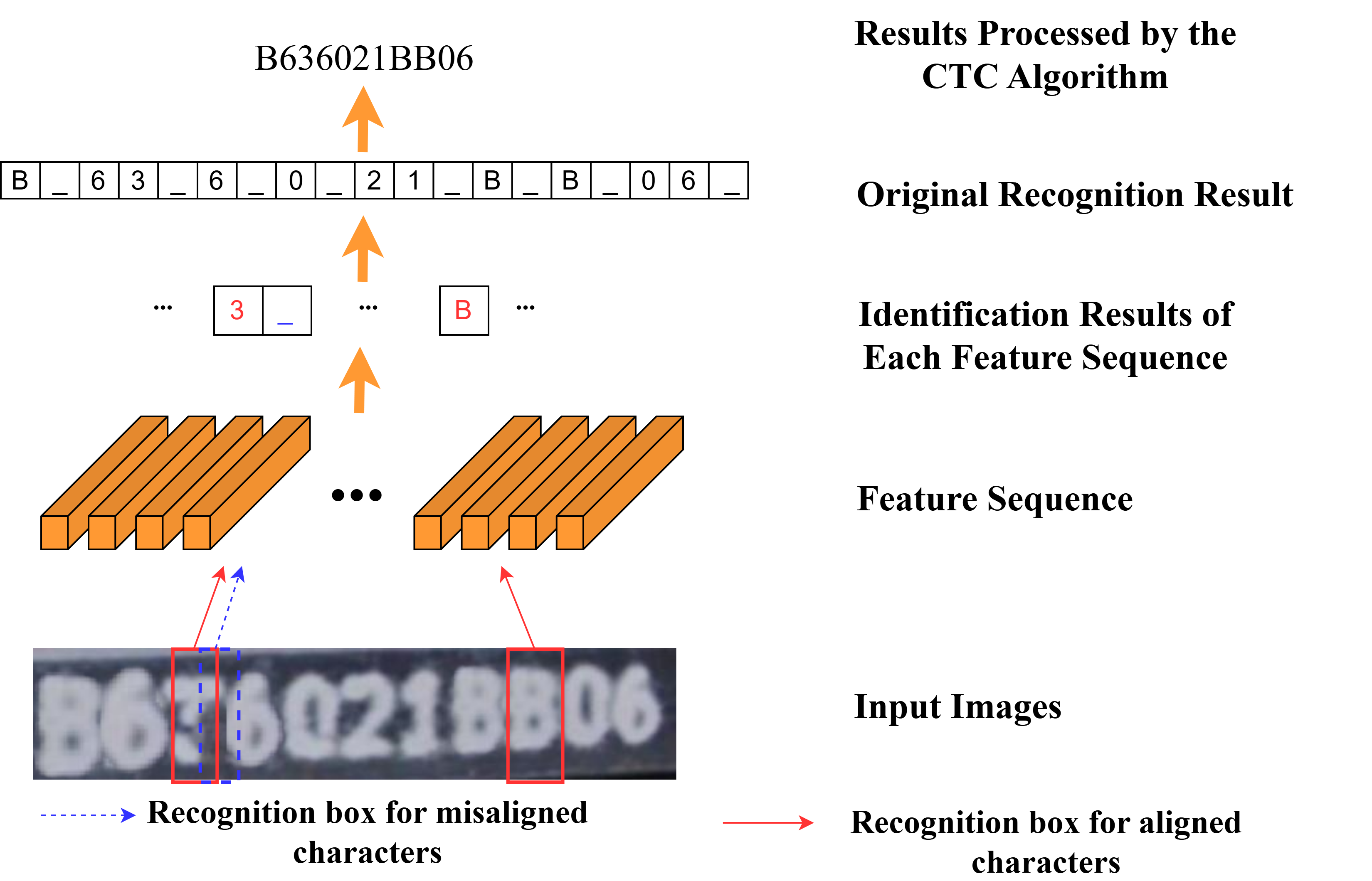}
   \caption{Architecture of the CTC algorithm.}
   \label{fig:onecol}
\end{figure}

In the task of steel billet printing mark recognition, the spray marks may suffer from various external influences such as damage or paint peeling. It has been observed that the CTC algorithm has certain limitations when dealing with damaged characters. Specifically, the CTC algorithm cannot distinguish whether a missing character is due to misalignment of the detection box or damage to the spray mark itself.

As illustrated in Figure 5, the left image shows a detection box aligned with the target character but with the character damaged, while the right image shows a detection box misaligned with the target character. In both cases, the recognition result will be labeled as ``\_" and subsequently removed during post-processing. While this behavior is reasonable for the second scenario, it results in missing recognition for the first scenario.

To address this issue, we incorporated a judgment mechanism into the CTC algorithm using prior knowledge. By analyzing the billet number images, we observed that the characters are closely spaced. By setting an appropriate image size, most detection boxes can be correctly aligned with the target characters. Under these circumstances, if three or more consecutive ``\_" labels appear, it can be inferred that a damaged character exists in that position. We then re-examine the recognition probabilities at that location and select the result with the highest probability, excluding the ``blank" label, as the recognition result. This approach mitigates the potential errors in CTC recognition caused by damaged characters.

\begin{figure}[h]
  \centering
   \includegraphics[width=1\linewidth]{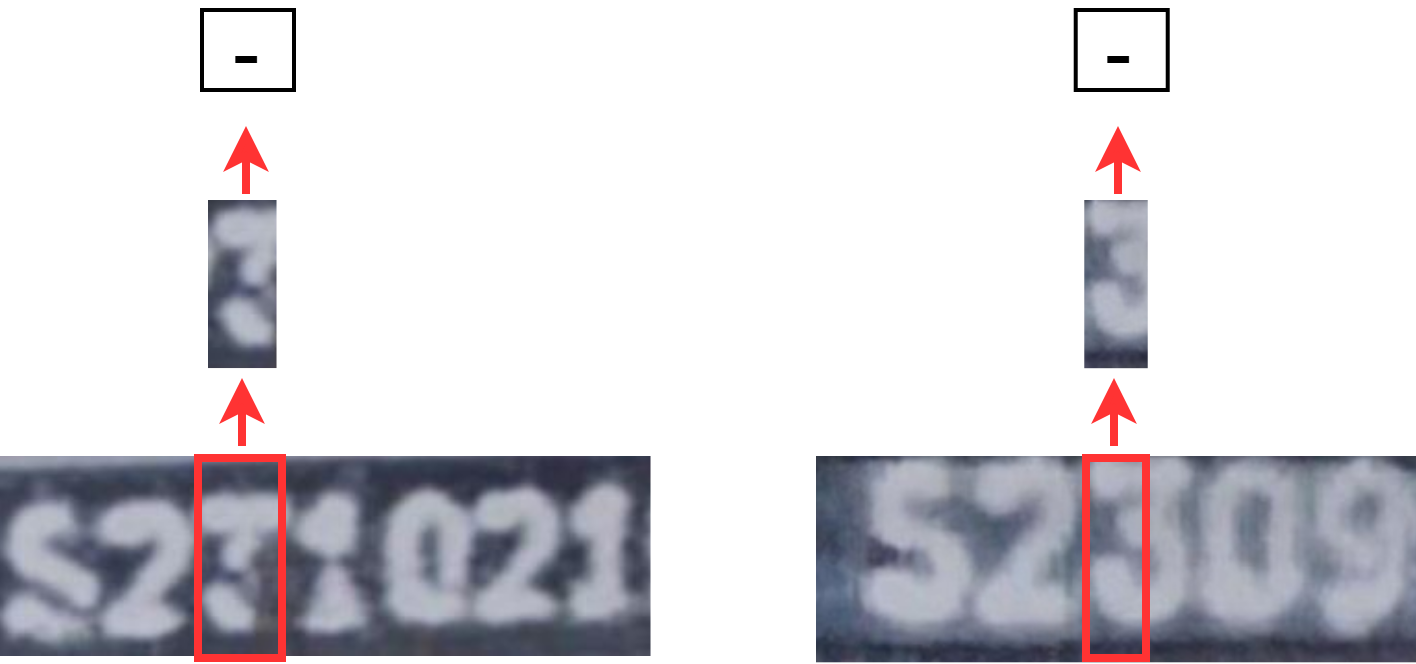}
   \caption{Misrecognition of the CTC algorithm in the case of damaged characters. The left image shows the case where the recognition box is aligned with the target character, but the character is damaged. The right image shows the case where the recognition box is not aligned with the target character. It can be seen that both cases yield the same recognition result.}
   \label{fig:onecol}
\end{figure}

\section{Experiment}
\subsection{Dataset}

We collected 4,725 machine-printed number images for fine-tuning the SVTR network. Among these, 4,200 images were used as the training set for fine-tuning, and 525 images were used as the validation set.

During the inference phase, we used an additional 500 machine-printed number images and 500 handwritten billet number images as the test set. 
% Some sample images from the machine-sprayed dataset are shown in \textbf{Figure 7(a)}, while samples from the handwritten billet number dataset are displayed in \textbf{Figure 7(b)}.

%\begin{figure*}
%  \centering
%  \begin{subfigure}{0.49\linewidth}
%    \includegraphics[width=\linewidth]{pic/组合1.png} % 替换为你的图片路径
%    \caption{Images of billet numbers printed by machine.}
%    \label{fig:short-a}
%  \end{subfigure}
%  \hfill
%  \begin{subfigure}{0.49\linewidth}
%    \includegraphics[width=\linewidth]{pic/组合2.png} % 替换为你的图片路径
%    \caption{Images of billet numbers printed by handwritten.}
%    \label{fig:short-b}
%  \end{subfigure}
%  \caption{Images of billet numbers datasets.}
%  \label{fig:short}
%\end{figure*}

\subsection{Experiment Settings}

The implementation of the DB and SVTR networks referenced the PaddleOCR\cite{25}. Model fine-tuning was conducted on an RTX 3090 GPU using a PyTorch-based deep learning framework on an Ubuntu 18.04 system. We employ ResNet50 as the backbone network. The parameters for the DB network are detailed in Table 2, while those used during the fine-tuning of the SVTR network are presented in Table 3.

\begin{table}
  \centering
  \caption{DB network parameter settings.}
  \label{tab:example}
  \begin{tabular}{@{}lc@{}}
    \toprule
    Parameters & Value \\
    \midrule
    det\_db\_thresh & 0.3 \\
    det\_db\_box\_thresh & 0.6\\
    det\_db\_unclip\_ratio & 1.5\\
    det\_limit\_side\_len & 2500\\
    use\_dilation & false\\
    \bottomrule
  \end{tabular}
\end{table}

\begin{table}
  \centering
  \caption{SVTR network parameter settings.}
  \label{tab:example}
  \begin{tabular}{@{}lc@{}}
    \toprule
    Parameters & Value \\
    \midrule
    learning rate & 0.005 \\
    batch size & 256 \\ 
    optimizer & adam\\
    beta1 & 0.9\\
    beta2 & 0.99\\
    regularization & L2\\
    drop score & 0.5\\
    warm up epoch & 5\\
    \bottomrule
  \end{tabular}
\end{table}

We used accuracy and edit distance as the metrics to evaluate the recognition performance of the model. For each billet number image, we consider the recognition correct only if every character in the image is correctly identified. Accuracy is calculated as the percentage of correctly recognized images relative to the total number of images in the test set. Edit distance refers to the minimum number of editing operations required to transform one string into another. The smaller the edit distance, the more similar the content of the two strings.

\subsection{Experiment Results}

\begin{table*}[!ht]
  \centering
  \caption{Algorithm performance evaluation on the machine-printed number test set.}
  \label{tab:twocolumn}
  \begin{tabular}{@{}lcc@{}}
    \toprule
    Method & Accuracy & Eidt distance\\
    \midrule
    DB + SVTR & 0.5841 & 1.2024 \\
    DB + SVTR + Inference with Prior Knowledge Constraints & 0.6852 & 0.7772 \\
    DB + SVTR + TTA & 0.7637 & 0.5835\\
    DB + SVTR + TTA + Inference with Prior Knowledge Constraints & 0.7990 & 0.5617\\
    \bottomrule
  \end{tabular}
\end{table*}

\begin{table*}[!ht]
  \centering
  \caption{Algorithm performance evaluation on the handwritten number test set.}
  \label{tab:twocolumn}
  \begin{tabular}{@{}lcc@{}}
    \toprule
    Method & Accuracy & Eidt distance\\
    \midrule
    DB + SVTR & 0.2371 & 4.3564 \\
    DB + SVTR + Fine-tuning & 0.5634 & 2.0534 \\
    DB + SVTR + Fine-tuning + Inference with Prior Knowledge Constraints & 0.6356 & 1.2958 \\
    DB + SVTR + Fine-tuning + TTA & 0.6862 & 1.0585\\
    DB + SVTR + Fine-tuning + TTA + Inference with Prior Knowledge Constraints & 0.7043 & 1.0313\\
    \bottomrule
  \end{tabular}
\end{table*}

Table 4 shows the inference results of using the DB network combined with an unrefined SVTR network as the baseline on the machine-printed billet number test set. The results indicate that the model performs poorly on this dataset, achieving an accuracy of only 0.58 and an average edit distance of 1.2, which are insufficient to meet practical production requirements. The primary reasons for the low recognition accuracy include the complex backgrounds of the billet number images, which introduce significant interference, and other objective factors such as variations in lighting, character positioning, and the quality of character printing. These challenges further hinder the model's recognition capability. Furthermore, although the baseline network had undergone extensive pretraining on a large dataset, the distribution mismatch between training and test data caused a domain shift, which adversely affected the model’s inference accuracy.

To address these issues, we introduced a prior knowledge-constrained inference method that allows the model to automatically correct predictions based on encoding rules. This approach also mitigates the recognition errors caused by damaged characters in the CTC module, leading to an improvement in accuracy to 0.68 and a reduction in the average edit distance to 0.77. Subsequently, by incorporating test-time adaptation (TTA), the model adapted its parameters during the testing phase to align better with the distribution of unseen images, thereby overcoming the challenges posed by domain shift. As a result, the recognition accuracy improved to 0.76, and the average edit distance decreased to 0.58. Finally, by combining both methods, the model achieved further performance gains, with the accuracy reaching 0.79 and the average edit distance reducing to 0.56. These results demonstrate the effectiveness of the proposed methods in addressing the challenges mentioned above.

Table 5 presents the experimental results on the handwritten billet number test set. Compared to machine-printed billet numbers, handwritten billet numbers exhibit poorer character quality, including issues such as significant character adhesion and irregular writing styles. As a result, the baseline model achieves only 0.23 recognition accuracy, with an average edit distance of 4.36. Due to the scarcity of handwritten samples, it is challenging to construct a targeted training set for supervised fine-tuning. After fine-tuning on the machine-printed billet number training set, the recognition accuracy improves to 0.56, and the average edit distance decreases to 2.05, as machine-printed images share certain similarities with handwritten billet images. However, the domain shift issue remains significant. By introducing a prior knowledge-constrained inference method, the model can automatically validate and correct some recognition errors, increasing the accuracy to 0.64 and reducing the edit distance to 1.29. Furthermore, the incorporation of test-time adaptation alleviates the domain shift between machine-printed and handwritten billet numbers, further improving the accuracy to 0.68 and reducing the average edit distance to 1.05. Finally, when both methods are applied to the fine-tuned model, the recognition accuracy and edit distance for handwritten billet numbers show substantial improvement, reaching 0.7 and 1.03, respectively. These experimental results demonstrate that the two proposed methods effectively enhance the performance of the pre-trained text recognition model on this dataset, even when fine-tuning for handwritten samples is difficult.

\section{Conclusion}

In this paper, we introduce a model for recognizing billet numbers that integrates the DB network for text detection and the SVTR network for text recognition to handle billet number identification. To address the challenge of domain shift, we incorporate a test-time adaptation technique into the billet number recognition model. By minimizing entropy during testing, the model adjusts to the distribution of billet images in actual production settings, achieving satisfactory recognition accuracy even in the absence of ample data and labels for fine-tuning. Furthermore, we utilize the encoding principles of billet numbers and rectify character recognition errors due to damaged characters in the CTC algorithm by introducing a prior knowledge-constrained inference approach. Experiments carried out on billet number dataset illustrate the significant enhancement in recognition accuracy in real-world production scenarios through these methods.

Nonetheless, this research has some limitations. The incorporation of TTA unavoidably leads to an increase in computational load. Also, the existence of noisy data in the test set might prompt the model to overfit to noise or outliers, thus diminishing its overall performance in generalization. Subsequent efforts will concentrate on refining the TTA algorithm to better align with the demands of the billet number recognition task.

%%%%%%%%% REFERENCES
{\small
\bibliographystyle{unsrt}
\bibliography{egbib}
}

\end{document}